\documentclass[letterpaper, 10 pt, conference]{ieeeconf}  

\IEEEoverridecommandlockouts                              
\usepackage[utf8]{inputenc}
\usepackage[T1]{fontenc}
\usepackage[disable, textsize=scriptsize]{todonotes}
\usepackage{lipsum}
\usepackage{algorithm}
\usepackage{algpseudocode}
\usepackage{url}
\usepackage{blindtext}
\usepackage[separate-uncertainty = true]{siunitx}
\usepackage{color,soul}
\usepackage{amsmath} 
\usepackage{amsfonts}
\usepackage{amssymb}  

\usepackage{shortcuts_ln} 

\algblock{ParFor}{EndParFor}
\algnewcommand\algorithmicparfor{\textbf{for}}
\algnewcommand\algorithmicpardo{\textbf{do parallel}}
\algnewcommand\algorithmicendparfor{\textbf{end\ for}}
\algrenewtext{ParFor}[1]{\algorithmicparfor\ #1\ \algorithmicpardo}
\algrenewtext{EndParFor}{\algorithmicendparfor}

\newcommand\pl[1]{}

\makeatletter
\newcommand\fs@betterruled{%
  \def\@fs@cfont{\bfseries}\let\@fs@capt\floatc@ruled
  \def\@fs@pre{\vspace*{5pt}\hrule height.8pt depth0pt \kern2pt}%
  \def\@fs@post{\kern2pt\hrule\relax}%
  \def\@fs@mid{\kern2pt\hrule\kern2pt}%
  \let\@fs@iftopcapt\iftrue}
\floatstyle{betterruled}
\restylefloat{algorithm}
\makeatother

\title{\LARGE \bf
Skill-based Multi-objective Reinforcement Learning of Industrial Robot Tasks with Planning and  Knowledge Integration
}

\author{Matthias Mayr$^{1}$, Faseeh Ahmad$^{1}$, Konstantinos Chatzilygeroudis$^{2}$, Luigi Nardi$^{1,3}$ and Volker Krueger$^{1}$
	\thanks{$^{1}$Department of Computer Science, Faculty of Engineering (LTH), Lund University, SE~221~00 Lund, Sweden. E-mail: <firstname>.<lastname>@cs.lth.se.
	}%
	\thanks{$^2$Computer Engineering and Informatics Department (CEID), University of Patras, Greece. E-mail: costashatz@upatras.gr.}
	\thanks{$^{3}$
	Department of Computer Science and Electrical Engineering, Stanford University, CA 94305, USA. E-mail: lnardi@stanford.edu.}
}

\begin{document}

\maketitle
\thispagestyle{empty}
\pagestyle{empty}

\begin{abstract}

In modern industrial settings with small batch sizes it should be easy to set up a robot system for a new task. Strategies exist, e.g. the use of skills, but when it comes to handling forces and torques, these systems often fall short.
We introduce an approach that provides a combination of task-level planning with targeted learning of scenario-specific parameters for skill-based systems.
We propose the following pipeline: the user provides a task goal in the planning language \textit{PDDL}, then a plan (i.e., a sequence of skills) is generated and the learnable parameters of the skills are automatically identified, and, finally,
an operator chooses reward functions and hyperparameters for the learning process.
Two aspects of our methodology are critical: (a) learning is tightly integrated with a knowledge framework to support symbolic planning and to provide priors for learning, (b) using
multi-objective optimization. This can help to balance key performance indicators (KPIs) such as safety and task performance since they can often affect each other. 
We adopt a multi-objective Bayesian optimization approach 
and learn entirely in simulation.
We demonstrate the efficacy and versatility of our approach by learning skill parameters for two different contact-rich tasks. We show their successful execution on a real 7-DOF \emph{KUKA-iiwa} manipulator and outperform the manual parameterization by human robot operators.

\end{abstract}

\section{Introduction}
Industrial environments with expensive and fragile equipment, safety regulations and frequently changing tasks often have special requirements for the behaviour policies that control a robot:
First, the trend in industrial manufacturing is to move to smaller batch sizes and higher flexibility of work stations. Reconfiguration needs to be fast, easy and should minimize downtime.
Second, it is important to be able to guarantee the performance as well as safety for material and workers. Therefore, it is crucial to be able to understand  \emph{what} action is performed \emph{when} and \emph{why}.
Finally, in industrial environments digital twins provide a lot of task-relevant information such as material properties and approximate part locations that the robot behavior policies have to consider.

One way to fulfill these criteria is to use systems based on parameterized \textit{skills}~\cite{krueger19rcim,krueger16ieee, bogh2012does}. These encapsulated abilities realize semantically defined actions such as moving the robot arm, opening a gripper or localizing an object with vision. State-of-the-art skill-based software architectures can not only utilize knowledge, but also automatically generate plans (skill-sequences) for a given task~\cite{crosby17icaps,rovida16icaps}. The skill-based approach is powerful when knowledge can be modeled and formalized \emph{explicitly}~\cite{krueger19rcim,krueger16ieee}. But it is often limited when it comes to skill parameters of contact-rich tasks that are difficult to reason about. One example are the parameters of a peg insertion search strategy where material properties (e.g. friction) and the robot behavior need to be considered. While it is possible to create a reasoner that follows a set of rules to determine such skill parameters, it is challenging to implement and to maintain.

Another way to handle this is to have operators manually specify and try values for these skill parameters. However, this is a manual process and can be cumbersome.

%
\begin{figure}[tpb]
	{
		\setlength{\fboxrule}{0pt}
		\framebox{\parbox{3in}{
		\includegraphics[width=0.95\columnwidth]{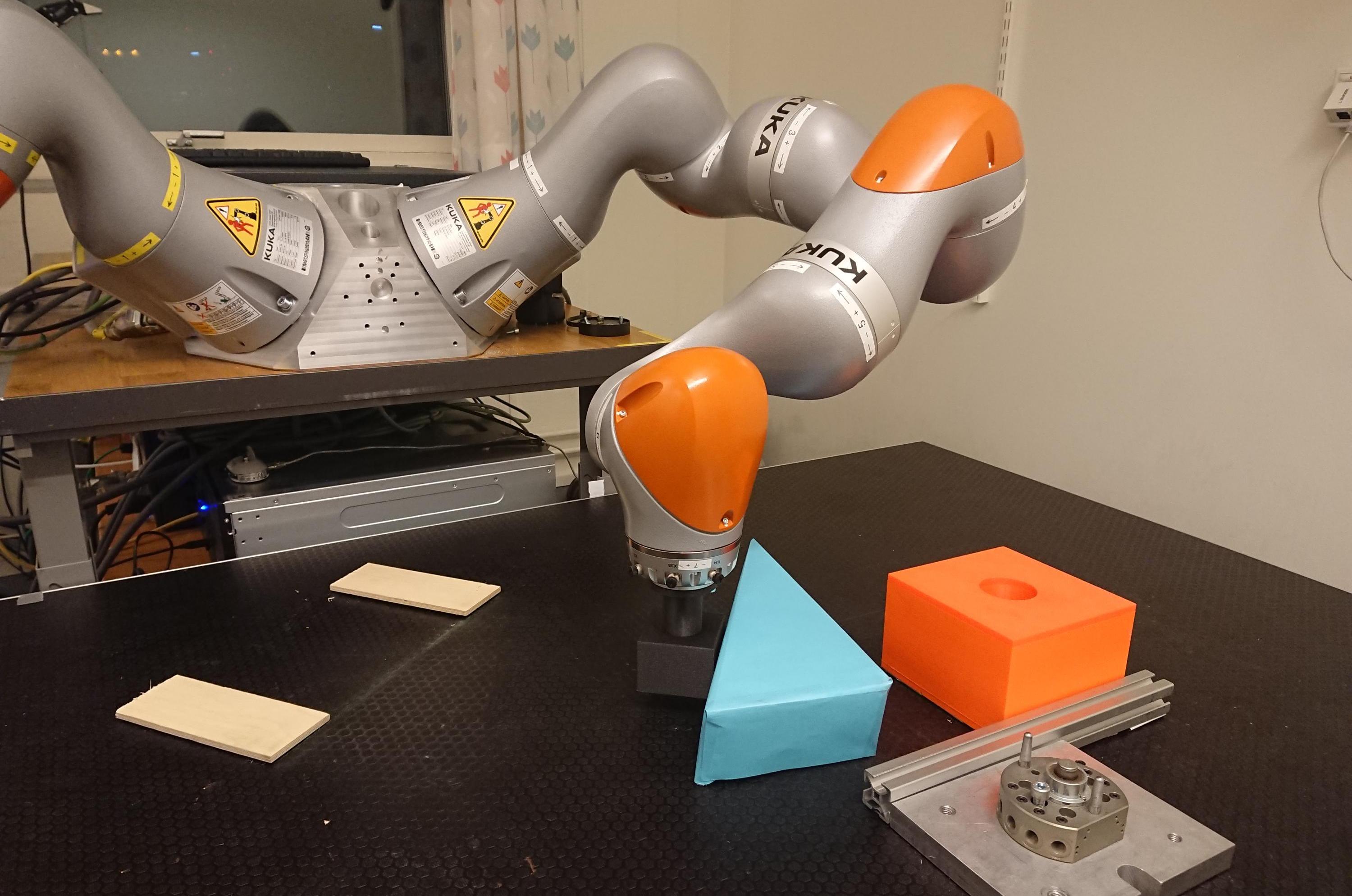}}}
	}
	\caption{The robot setup used for the experiments. Wooden boards indicate the start location for the push task. The goal is the corner between the fixture and the box with the hole for the peg task.}
	\label{fig:robot_setup}
	\vspace{-2em}
\end{figure}
%
Finally, it is possible to allow the system to learn by interacting with the environment. However, many policy formulations that allow learning (e.g. artificial neural networks) have deficiencies which make their application in an industrial domain with the abovementioned requirements challenging. Primarily during the learning phase, dangerous behaviors can be produced and even state-of-the-art RL methods need hundreds of hours of interaction time~\cite{chatzilygeroudis2019survey}. Learning in simulation can help to reduce downtime and dangers for the real system. But many policy formulations are black boxes for operators and it can be hard to predict their behavior, which could hinder the trust to the system~\cite{edmonds2019tale} Moreover, the simulation-to-reality gap~\cite{koos2012transferability,mouret201720} is bigger in lower-level control states (i.e. torques), and policies working directly on raw control states struggle to transfer learned behaviors to the real systems~\cite{chatzilygeroudis2019survey}. Our policy formulation consisting of behavior trees (BT) with a motion generator~\cite{rovida182iicirsi} has shown to be able to learn interpretable and robust behaviors~\cite{mayr21iros}.

The formulation of a learning problem for a given task is often not easy and becomes more challenging if factors such as safety or impact on the workstation environment need to be considered.
 Multi-objective optimization techniques allow to specify multiple objectives and optimize for them concurrently. This allows operators to
 select from solutions that are optimal for a certain trade-off between the objectives (usually represented as a set of Pareto-optimal solutions).
In order to learn sample-efficient and to support the large variety of skill implementations as well as scenarios, we use gradient-free Bayesian optimization as an optimization method.
%

\noindent In this paper we make the following contributions:

\begin{enumerate}
    \item We introduce a new method which seamlessly integrates symbolic planning and reinforcement learning for skill-based systems 
    to learn interpretable policies for a given task.
    \item A Bayesian multi-objective treatment of the task learning problem,  which includes the operator through easy specification of problem constraints and task objectives (KPIs); the set of Pareto-optimal solutions is presented to the operator and their behavior can be inspected in simulation and executed on the real system.
    \item We demonstrate our approach on two contact-rich tasks, a pushing task and a peg-in-hole task. We compare it to the outcome of the planner without reasoning, 
    randomly sampled parameter sets from the search space 
    and the manual real-world parameterization process of robot operators. In both tasks our approach delivered solutions that even outperform the ones found by the manual search of human robot operators.
\end{enumerate}

\section{Related Work}
\subsection{Skill-based Systems}
Skill-based systems are one way to support a quick setup of a robot system for a new task and to allow re-use of capabilities. There are multiple definitions of the term \emph{skills} in the literature. Some define it as a pure \emph{motion skills}~\cite{hasegawa1991model} or "hybrid motions or tool operations"~\cite{thomas2003error}. Other work has a broader skill definition~\cite{krueger19rcim, krueger16ieee, bogh2012does, crosby17icaps, rovida16icaps, rovida2017skiros, thomas2013new}. In this formulation, skills can be arbitrary capabilities that change the state of the world and have pre- and postconditions. Their implementation can include motion skills, but also proficiencies such as vision-based localization of objects. In~\cite{stenmark2015distributed} skills are "high-level reusable robot capabilities, with the goal to reduce the complexity and time consumption of robot programming". However, compared to~\cite{bogh2012does} and~\cite{rovida2017skiros} they do not use pre- and postconditions.
In~\cite{7989070}, an integrated system for manual creation of \emph{task plans} is presented. It shares the usage of BTs with our approach. 

Task planners are used in~\cite{krueger19rcim, krueger16ieee, crosby17icaps, rovida16icaps, rovida2017skiros, rovida2017extended, stenmark2015distributed, thomas2003error} while~\cite{7989070} lacks such a capability.

In~\cite{stenmark2015distributed} it is suggested that "Machine learning can be performed on the motion level, in terms of adaptation, or can take the form of structured learning on a task/error specification level". However, none of the reviewed work offers a combination task-level planning with learning.

\subsection{Policy Representation and Learning}
An important decision to make when working with manipulators is the type of policy representation and on which level it interfaces with the robot. The latter can strongly influence the learning speed and the quality of the obtained solutions~\cite{varin2019policy, martin2019imp}. These choices also influence the form of priors that can be defined and how they are defined~\cite{chatzilygeroudis2019survey}.
Not many policies combine the aforementioned properties of being a) interpretable, b) paramterizable for the task at hand and c) allow learning or improvement.

The commonly used policy representations for learning systems include radial basis function networks~\cite{deisenroth13r}, dynamical movement primitives~\cite{ijspeert2013dynamical, ude2010task} and feed-forward neural networks~\cite{deisenroth13r,chatzilygeroudis172iicirsi}. In recent years deep artificial neural networks (ANN) seem to become a popular policy.
All of them have in common that their final representation can be difficult to interpret. Even if a policy only sets a target pose for the robot to reach, it can be problematic to know how it reacts in all parts of the state space. %
In contrast to that,~\cite{mayr21iros} suggests to learn interpretable policies based on behavior trees~\cite{rovida182iicirsi} that work explicitly in end-effector space.
\subsection{Planning and Learning}
Symbolic planning is combined with learning in \cite{grounds2005combining,gordon2019should,yang2018peorl,sarathy2020spotter}. In ~\cite{grounds2005combining}, the PLANQ-Learning algorithm uses a symbolic planner to shape the reward function based on the conditions defined which are then used by the Q-learner to get an optimal policy with good results on the grid domain. \cite{gordon2019should} uses the combined symbolic planner with reinforcement learning (RL) in a hierarchical framework to solve complex visual interactive question answering tasks. PEORL~\cite{yang2018peorl} integrates symbolic planning and  hierarchical reinforcement learning (HRL) to improve performance by achieving rapid policy search and robust symbolic planning in the taxi domain and grid world. SPOTTER~\cite{sarathy2020spotter} uses RL to allow the planning agent to discover the new operators required to complete tasks in Grid World. In contrast to all these approaches, our approach aims towards real-life robotic tasks in an Industry 4.0 setting where a digital twin is available.

In~\cite{styrud2021combining},  the authors combine symbolic planning with behavior trees (BT) to solve blocks world tasks with a robot manipulator. They use modified Genetic Programming (GP)~\cite{koza1992genetic} to learn the structure of the BT. In our approach, we focus on learning the parameters of the skills in the BT and utilize a symbolic planner to obtain the structure of the BT.

\section{Approach}

Our approach consists of two main components that interact in different stages of the learning pipeline: First, \textit{SkiROS}~\cite{rovida2017skiros}, a skill-based framework for ROS, which represents the implemented skills with BTs, hosts the world model (digital twin), and interacts with the planner. \textit{SkiROS} is also used to execute BTs while learning and to perform tasks on the real system. Second, the learning framework that provides the simulation, the integration with the policy optimizer as well as the reward function definition and calculation.
\begin{figure}[tpb]
	{
		\setlength{\fboxrule}{0pt}
		\framebox{\parbox{3in}{
		\includegraphics[width=0.92\columnwidth]{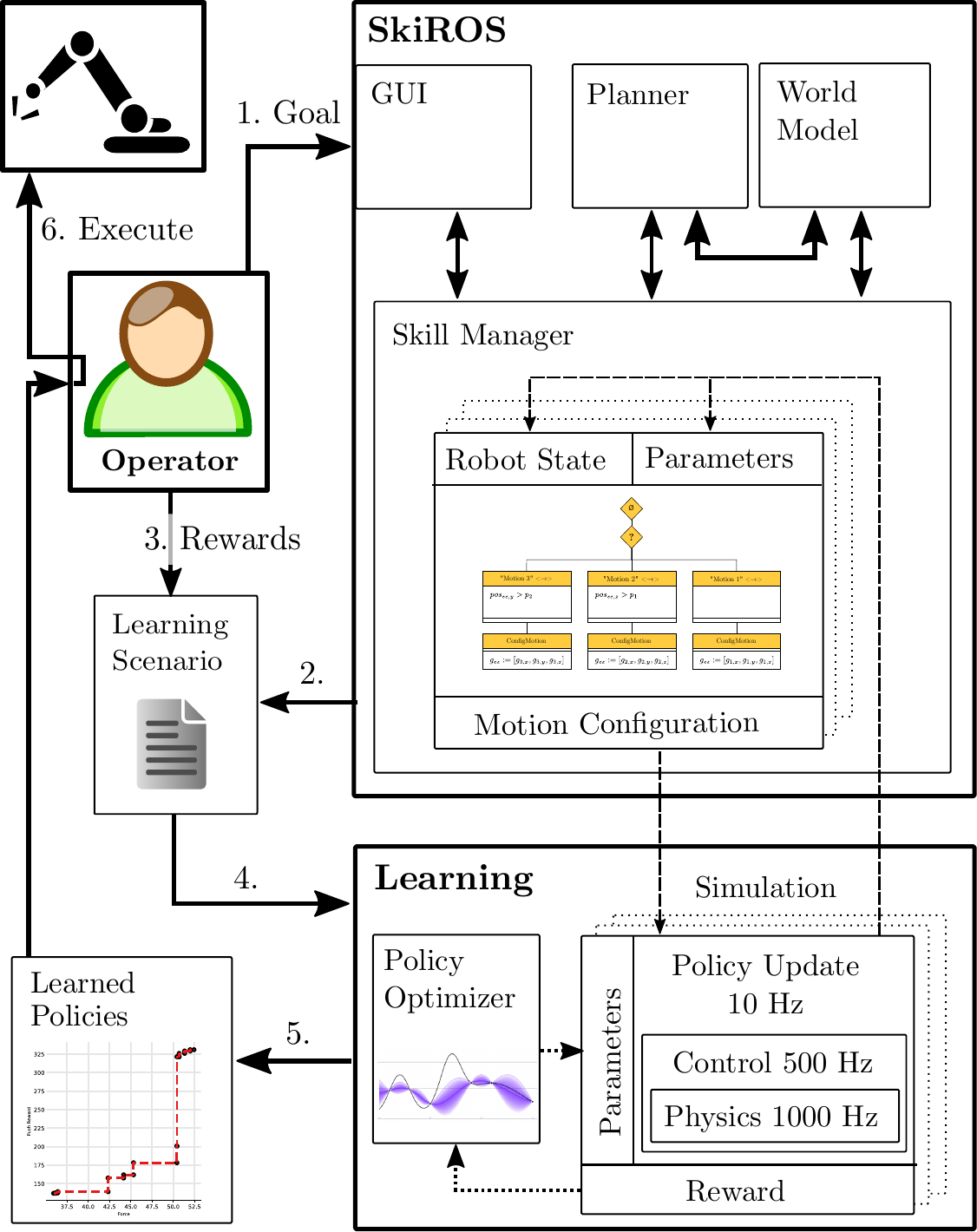}}}
	}
	\caption{The architecture of the system that depicts the pipeline: (1) The operator enters the goal state; (2) a learning scenario for the plan is created; (3) rewards and hyperparameters are specified; (4) learning is conducted using the skills and the information in the world model; (5) after policy learning, the operator can choose which policies to execute on the real system (6).}
	\label{fig:system}
	\vspace{-1em}
\end{figure}
The architecture of the system and the workflow is shown in Figure~\ref{fig:system}: (1) an operator enters the task goal into a GUI; (2) a plan with the respective learning scenario configuration is generated; (3) an operator complements the scenario with objectives and reward functions; (4) learning is conducted in simulation using the skills and information from the world model; (5) in the multi-objective optimization case, a set of Pareto-optimal solutions is generated and presented to the operator; finally, (6) the operator can select a good solution from this set given the desired trade-off between KPIs and execute it on the real system.
%
\subsection{Behavior Trees}
A Behavior Tree (BT)~\cite{colledanchise17ac}\todo{add BT survey} is a formalism for plan representation and execution.
Like~\cite{rovida172iicirsi,marzinotto142iicrai}, we define it as a directed acyclic graph $G(V,W)$ with $|V|$ nodes and $|W|$ edges. It consists of \emph{control flow nodes} (\emph{processors}), and \emph{execution nodes}.
The four basic types of \emph{control flow nodes} are 1)~\emph{sequence},  2)~\emph{selector}, 3)~\emph{parallel} and 4)~\emph{decorator}~\cite{marzinotto142iicrai}.
A BT always has one initial node with no parents, defined as \emph{Root}, and one or more nodes with no children, called \emph{leaves}.
When executing a BT, the \emph{Root} node periodically injects a \emph{tick} signal into the tree. The signal is routed through the branches according to the  implementation of the \emph{control flow nodes} and the return statements of their children. By convention, the signal propagation goes from left to right.

\begin{figure}[tpb]
	{
		\setlength{\fboxrule}{0pt}
				\framebox{\parbox{3in}{
				\includegraphics[width=0.95\columnwidth]{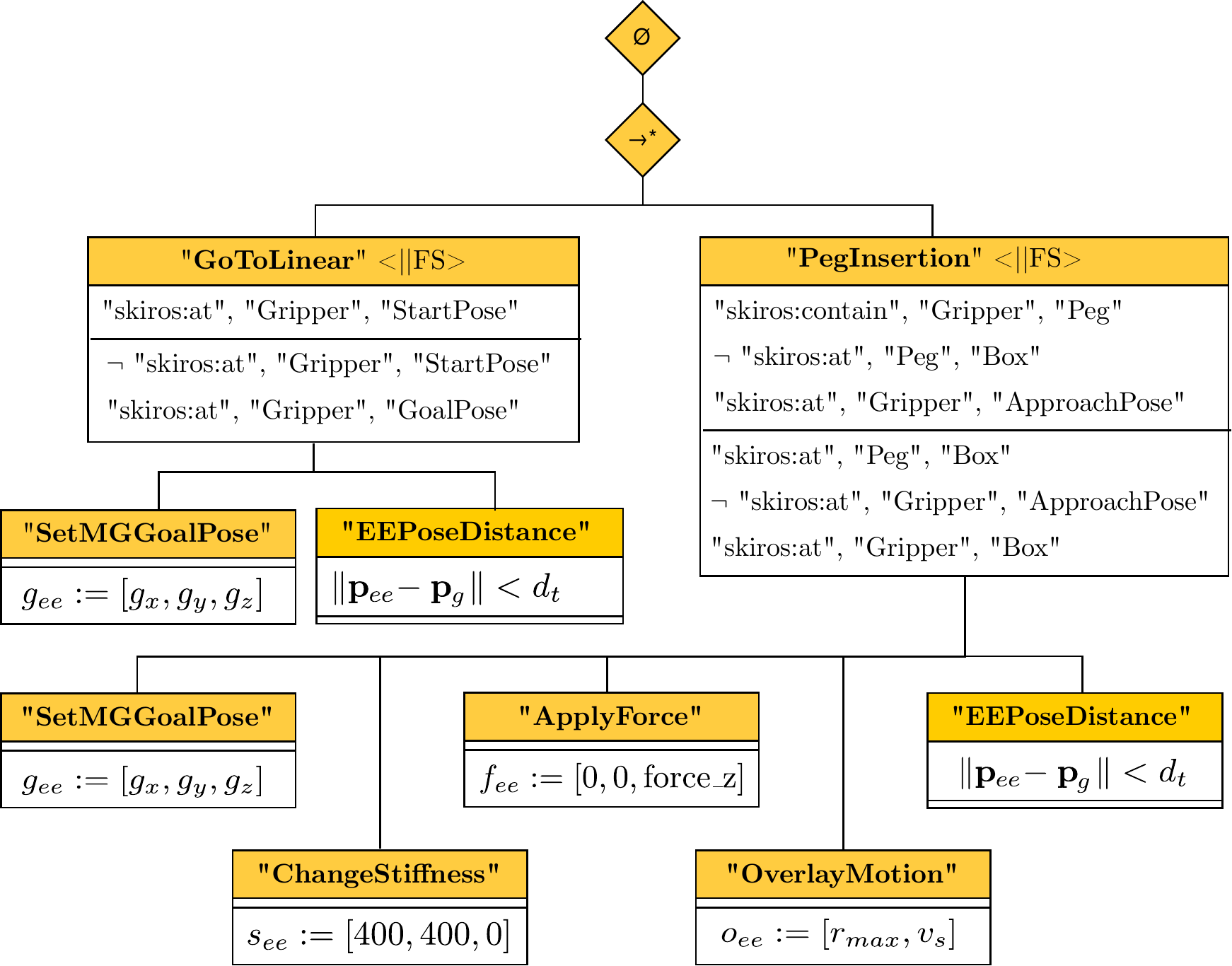}}}
	}
	\caption{The BT of the generated plan for the peg insertion task in \textit{eBT format}~\cite{rovida172iicirsi}. Each node has conditions or pre-conditions shown in the upper half and effects or post-conditions shown in the lower half. The \emph{serial start control flow node} (\(\rightarrow^*\)) executes in a sequence and remembers the successes. The skills have a \textit{parallel-first-success} processor (<\(||FS\)>).}
	\label{fig:behavior_tree}
	\vspace{-1em}
\end{figure}

The \emph{sequence} node corresponds to a logical \emph{AND}: it succeeds if all children succeed and fails if one child fails. The \emph{selector}, also called \emph{fallback} node, represents a logical \emph{OR}: 
If one child succeeds, the remaining ones will not be ticked. It fails only if all children fail. The \emph{parallel} control flow node forwards ticks to all children and fails if one fails. A \emph{decorator} allows to define custom functions. Implementations like \textit{extended Behavior Trees} (eBT) in \textit{SkiROS}~\cite{rovida172iicirsi} add custom processors such as \emph{parallel-first-success} that succeeds if one of the parallel running children succeeds.
Leaves of the BT are the \emph{execution nodes} that, when ticked, execute one cycle and output one of the three signals: \emph{success}, \emph{failure} or \emph{running}. In particular, execution nodes subdivide into 1)~\emph{action} and 2)~\emph{condition} nodes. An action performs its operation iteratively at every tick, returning \emph{running} while it is not done, and \emph{success} or \emph{failure} otherwise. A condition performs an instantaneous operation and returns always \emph{success} or \emph{failure} and never \emph{running}. 
An example of the BT for the peg insertion task is in Fig.~\ref{fig:behavior_tree}. 

\subsection{Planning and Knowledge Integration}

The Planning Domain Definition Language (PDDL)~\cite{fox2003pddl2,crosby17icaps} is used to formulate the planning problem. 
We use the \textit{SkiROS}~\cite{rovida2017extended} framework that
automatically translates a task into a PDDL planning problem by generating domain description and problem instance using the world model.
We then use the semantic world model (WM) from \textit{SkiROS} ~\cite{rovida2017skiros} as the knowledge integration framework. 

Actions and fluents are obtained by utilizing the predicates that have pre- or post-conditions in the world model.
For the problem instance, the objects (robots, arms, grippers, boxes, poses, etc.) in the scene and their initial states (as far as they are known) are used.
After getting the necessary domain description and the problem instance \textit{SkiROS} calls the planner. The goal of the planner is to return a sequence of skills that can achieve the goal conditions of the task. The individual skills are partially parameterized with explicit data from the WM. The WM is aware of the skill parameters that need to be learned for the task at hand and they are automatically identified in the skill sequence.

\subsection{Policy Optimization}
\label{sec:optimization}
In order to optimize for policy parameters, we adopt the policy search formulation~\cite{deisenroth13r,chatzilygeroudis2019survey,chatzilygeroudis172iicirsi}. %
We formulate a dynamical system in the form:
\begin{equation}
\label{formula:dynamics}
\mathbf{x}_{t+1}=\mathbf{x}_{t}+M(\mathbf{x}_{t}, \mathbf{u}_{t}, \boldsymbol{\phi}_R)
,
\end{equation}
with continuous-valued states $\mathbf{x} \in \mathbb{R}^E$ and actions $\mathbf{u} \in \mathbb{R}^U$.
The transition dynamics are modeled by a simulation of the robot and the environment $M(\mathbf{x}_t, \mathbf{u}_t, \boldsymbol{\phi}_R)$. They are influenced by the domain randomization parameters $\boldsymbol{\phi}_R$.

The goal is to find a policy $\pi, \mathbf{u} = \pi(\mathbf{x}|\boldsymbol{\theta})$ with policy parameters $\boldsymbol{\theta}$ such that we maximize  the expected long-term reward when executing the policy for $T$ time steps:
\begin{equation}
\label{formula:long_term_reward}
J(\boldsymbol{\theta}) =\mathbb{E} \left[\sum_{t=1}^{T} r(\mathbf{x}_{t},\mathbf{u}_{t}) | \boldsymbol{\theta}\right]
,
\end{equation}
where $r(\mathbf{x}_{t},\mathbf{u}_{t})$ is the immediate reward for being in state $\mathbf{x}$ and executing action $\mathbf{u}$ at time step $t$.
The discrete switching of branches in the BT and most skills are not differentiable. Therefore, we frame the optimization in Eq.~\eqref{formula:long_term_reward} as a black-box optimization and pursue the maximization of the reward function $J(\mathbf{\theta})$ only by using measurements of the function. The optimal reward function to solve the task is generally unknown, and a combination of reward functions is usually used. In the RL literature, this is usually done with a weighted average, that is, $r(\mathbf{x}_{t},\mathbf{u}_{t})=\sum_iw_ir_i(\mathbf{x}_{t},\mathbf{u}_{t})$. In this paper, we chose not to use a weighted average of reward functions that represent different objectives (as the optimal combination of weights cannot always be found~\cite{kaushik2018multi}), but optimize for all objectives concurrently (Sec.~\ref{sec:multi-objective}) using Bayesian Optimization.
\subsection{Bayesian Optimization} 
\label{sec:bo}
We consider the problem of finding a global minimizer (or maximizer) of an unknown (black-box) objective function $f$:
$\mathbf{s^*} \in \argmin_{\mathbf{s} \in \bbS}
f(\mathbf{s}),$ 
where $\bbS$ is some input design space of interest in $D$ dimensions.
The problem addressed in this paper is the optimization of a 
(possibly noisy) function $f:\bbS \rightarrow \bbR$ 
with lower and upper bounds on the problem variables.
The variables defining $\bbS$ can be real (continuous), integer, ordinal, and categorical as in~\cite{nardi18hypermapper}. We assume that the function $f$ is in general expensive to evaluate
and that the derivatives of $f$ are in general not available. 
The function $f$ is called black box because we cannot access other information than the output $y$ given an input value $\mathbf{s}$. 


This problem can be tackled using Bayesian Optimization (BO)~\cite{frazier2018tutorial}. 
BO approximates $\mathbf{s}^*$ with a sequence of evaluations, $y_{1}, y_{2},\ldots, y_t$ at  $\mathbf{s}_1, \mathbf{s}_2, \ldots, \mathbf{s}_t \in \bbS$, which maximizes an utility metric, with each new $\mathbf{s}_{t+1}$ depending on the previous function values. BO achieves this by building a probabilistic surrogate model on $f$ based on the set of evaluated points $\{(\mathbf{s}_i, y_i)\}_{i=1}^{t}$. At each iteration, a new point is selected and evaluated based on the surrogate model which is then updated to include the new point $(\mathbf{s}_{t+1}, y_{t+1})$. 
BO defines an utility metric called the acquisition function, which gives a score to each $\mathbf{s} \in \bbS$ by balancing the predicted value and the uncertainty of the prediction for $\mathbf{s}$. 
The maximization of the acquisition function guides the sequential decision making process and the exploration versus exploitation trade-off: the highest score identifies the next point $\mathbf{s}_{t+1}$ to evaluate. 
BO is a statistically efficient black-box optimization approach when considering the number of necessary function evaluations~\cite{brochu2010tutorial}. It is, thus, especially well-suited to solve problems where we can only perform a limited number of function evaluations, such as the ones found in robotics. 

We use the implementation of BO found in \textit{HyperMapper}~\cite{nardi18hypermapper,nardi2017algorithmic,bodin2016integrating,souza2021bayesian}.
Our implementation selects the Expected Improvement (EI) acquisition function~\cite{mockus1978application} and we use uniform random samples as a warm-up strategy before starting the optimization.

\subsection{Multi-objective Optimization}
\label{sec:multi-objective}
Let us consider a multiple objectives minimization (or maximization) over $\bbS$ in $D$ dimensions.
We define $f:\bbS \rightarrow \mathbb{R}^{p}$ as our vector of objective functions $f = (f_{1}, \dots, f_{p})$,
taking $\mathbf{s}$ as input, and evaluating $y = f(\mathbf{s}) + \epsilon$, where $\epsilon$ is a Gaussian noise term. Our goal is to identify the Pareto frontier of $f$, that is,
the set $\Gamma \subseteq \bbS$ of points which are not dominated by any other point,
\ie the maximally desirable $\mathbf{s}$ which cannot be optimized further for any single objective without making a trade-off.
Formally, we consider the partial order in $\mathbb{R}^{p}$: $y \prec y'$ iff $\forall i \in [p], y_{i} \leqslant y'_{i}$
and $\exists j, \, y_{j} \! < \! y'_{j}$, and define the induced order on $\bbS$: $\mathbf{s} \prec \mathbf{s}'$ iff $f(\mathbf{s}) \prec f(\mathbf{s}')$.
The set of minimal points in this order is the Pareto-optimal set $\Gamma = \{\mathbf{s} \in \bbS : \nexists \mathbf{s}'$ such that $\mathbf{s}' \prec \mathbf{s}\}$. We aim to identify $\Gamma$ with the fewest possible function evaluations using BO. 
For this purpose we use the \emph{HyperMapper} multi-objective Bayesian optimization which is based on random scalarizations~\cite{paria18scalarizations}.

\subsection{Motion Generator and Robot Control}
\label{sec:control}
The arm motions are controlled in end-effector space by a Cartesian impedance controller.
The time varying \textit{reference} or \textit{attractor point} of the end effector $\mathbf{x}_d$ is governed by a motion generator (MG). 
Given the joint configuration \(\mathbf{q}\), we can calculate the end-effector pose \(\mathbf{x}_{ee}\) using forward kinematics and obtain an error term \(\mathbf{x}_e = \mathbf{x}_{ee} - \mathbf{x}_d\).
Together with the joint velocities \(\dot{\mathbf{q}}\), the Jacobian \(\mathbf{J}(\mathbf{q})\), the configurable stiffness and damping matrices \(K_{d}\) and \(D_{d}\), the task control is formulated as
$\mathbf{\tau}_{c}=\mathbf{J}^{T}(\mathbf{q})\left(-\mathbf{K}_{d} \mathbf{x}_{e}-\mathbf{D}_{d} \mathbf{J}(\mathbf{q}) \mathbf{\dot{q}}\right).$
Additionally, the task control can be overlayed with commanded generalized forces and torques \(\mathbf{F}_{ext}=\left(f_{x}~f_{y}~f_{z}~\tau_{x}~\tau_{y}~\tau_{z}\right)\):
$\mathbf{\tau}_{ext}=\mathbf{J}^{T}(\mathbf{q}) \mathbf{F}_{ext}.$
We utilize the integration introduced in~\cite{rovida182iicirsi} and used in~\cite{mayr21iros}, which proposes to parameterize the MG with movement skills from the BT. The reference pose is shaped by 1) a linear trajectory to a goal point and 2) overlay motions that can be added to the reference pose as discussed in~\cite{rovida182iicirsi, mayr21iros}. E.g. an Archimedes spiral for search.

To make it compliant with the dynamical system in Eq.~\eqref{formula:dynamics}, a new reference configuration of the controller is only generated at every time step $t$. It includes the reference pose, stiffnesses, applied wrench and forms the action \(\mathbf{u}\) with a dimension of \(U = 19\).  The stiffness and applied force are changed gradually at every time step \(t\) to ensure a smooth motion. The state space consists of joint positions and joint velocities and is \(E = 14\) dimensional.
Direct control of the torques of a robot arm requires high update rates and we control the robot arm at 500 Hz based on the current action \(\mathbf{u}\), but continuously
updated values for \(\mathbf{q}\) and \(\dot{\mathbf{q}}\). Therefore, from the perspective of Eq.~\eqref{formula:dynamics}, the controller is to be seen as part of the model \(M(\mathbf{x}_{t}, \mathbf{u}_{t})\).


We assume a human-robot collaborative workspace with fragile objects. Therefore, the stiffnesses and applied forces are to be kept to a minimum and less accuracy than e.g. high-gain position-controlled solutions is to be expected.
%
\section{Experiments}
\label{sec:experiments}
In our experiments we use a set of pre-defined skills that are part of a skill library. In order to solve a task, the planner determines a sequence that can achieve the goal condition of the task. This skill sequence is also automatically parameterized to the extend possible, e.g. the goal pose of a movement.
We evaluate our system in two contact-rich scenarios that are shown in  Fig. \ref{fig:robot_setup}: A) pushing an object with uneven weight distribution to a goal pose and B) inserting a peg in a hole with a \SI{1.5}{\milli\meter} larger radius. Pure planning-based solutions for both these tasks have a poor performance in reality~(Fig.~\ref{fig:success-results}).

As a baseline we invited six robot operators to manually parameterize the skills for the tasks. Their main objective is to find a parameter set that robustly solves the task. As an additional objective they were asked to minimize the impact of the robot arm and its tool on the environment as long as it does not affect the first objective.

The robot arm used for the physical evaluation is a 7-degree-of-freedom (DOF) \textit{KUKA iiwa} arm controlled by a Cartesian impedance controller (Sec.~\ref{sec:control}).


\subsection{Reward Functions}
\label{sec:rewards}

For each task, we utilize a set of reward functions parameterized for the learning scenario configuration. Each configured reward has an assigned objective and can be weighted against other rewards. Each experiment uses a subset of the following reward functions:
\subsubsection{Task completion} A fixed reward is assigned when the BT returns success upon task completion. 
\subsubsection{End-effector distance to a box} We use a localized reward to attract the end effector towards the goal location
$\label{formula:hole}
r_{h}(\mathbf{x})= \left(2\left(d(\mathbf{p}_{ee,\mathbf{x}}, \mathbf{p}_{h}) + d_o\right)\right)^{-1},$
where \(d_o\) is the distance offset and \(d(\mathbf{p}_{ee,\mathbf{x}}, \mathbf{p}_{h})\) is the shortest distance function  between the end effector and the box.\todo{M: Mention that this is used as an indicator for the force applied on the environment?}
\subsubsection{Applied wrench} This reward calculates the cumulative forces applied by the end effector on the environment.

Reward functions 4-6 share a common operation of computing an exponential function of the calculated metric to obtain the reward as used in (\cite{deisenroth11p2icicml, chatzilygeroudis172iicirsi}) 
$\label{formula:reward_exp}
r(d_m)=\exp \left(-\frac{1}{2 \sigma_{w}^{2}}(d_m + d_o)\right),$
where \(\sigma_w\) is a configurable width, \(d_o\) is a distance offset and \(d_m\) is the input metric.
\subsubsection{End-effector distance to a goal}This reward uses distance between the end effectors current pose and goal pose to calculate the input metric
$d_{ee,g} = \left\|\mathbf{p}_{ee,\mathbf{x}}-\mathbf{p}_{g}\right\|$
\subsubsection{End-effector-reference-position distance}This reward uses the distance between the end effectors reference pose~(Sec.~\ref{sec:control}) and its current pose to calculate the input metric
$d_{ee,d} = \left\|\mathbf{p}_{ee,\mathbf{x}}-\mathbf{x_d}\right\|$
\subsubsection{Object-pose divergence}This reward uses the translational and angular distance between the object's goal pose and its current pose.
%
%
\subsection{Push Task}
\label{sec:push-task}
The push task starts by specifying the goal in the \textit{SkiROS} Graphical User Interface (GUI) as: {\small\texttt{(skiros:at} \texttt{skiros:ObjectToBePushed-1}\texttt{ skiros:Object\-Goal\-Pose-1)}}. \textit{SkiROS} calls the planner to generate a plan given all the available skills. The plan consists of two skills: 1) $GoToLinear$ skill and 2) $Push$ skill. The first skill moves the end effector from its current location to the \emph{approach pose} of the object. This \emph{approach pose} is defined in the WM and needs to be reached before interacting with the object.

The push skill then moves the end effector to the object's geometric centre with an optional offset in the horizontal (\(x\)) and (\(y\)) directions. 
Once the end effector reaches it, the motion generator executes a straight line to the (modified) target location.

The push task is formulated as a multi-objective task. It also has two objectives, 1) success and 2) applied force. The first objective has three associated rewards: 1) object position difference from goal position, 2) object orientation difference from goal orientation, and 3) end-effector distance to the goal location. The second objective accumulates the Cartesian distance between the end-effector reference pose and the actual end-effector pose as a measure of the force applied by the controller.
The learnable parameters in this task are offsets in the horizontal (\(x\)) and (\(y\)) direction of both the push skill's start and goal locations. An offset of the start location allows the robot to push from a particular point from the side of the object. Together with the offsets on the goal position, these learnable parameters collectively define the trajectory of the push.

The object to be pushed has a height of \SI{0.07}{\meter} and is an orthogonal triangle in the horizontal dimensions (\(x\)) and (\(y\)). It has a length of \SI{0.15}{\meter} and \SI{0.3}{\meter} and it weights \SI{2.5}{\kilogram}. For this task we use a square-shaped peg for pushing with a side length of \SI{0.07}{\meter} and a height of \SI{0.05}{\meter}. Start and goal locations are \SI{\approx 0.43}{\meter} apart and are rotated by \(26~deg\). We define success if the translational and rotational difference of the object w.r.t the goal is less than \SI{0.01}{\meter} and \(5~deg\), respectively.
%
\begin{figure}[tpb]
	{
		\setlength{\fboxrule}{0pt}
		\framebox{\parbox{3in}{
				\includegraphics[width=0.9\columnwidth]{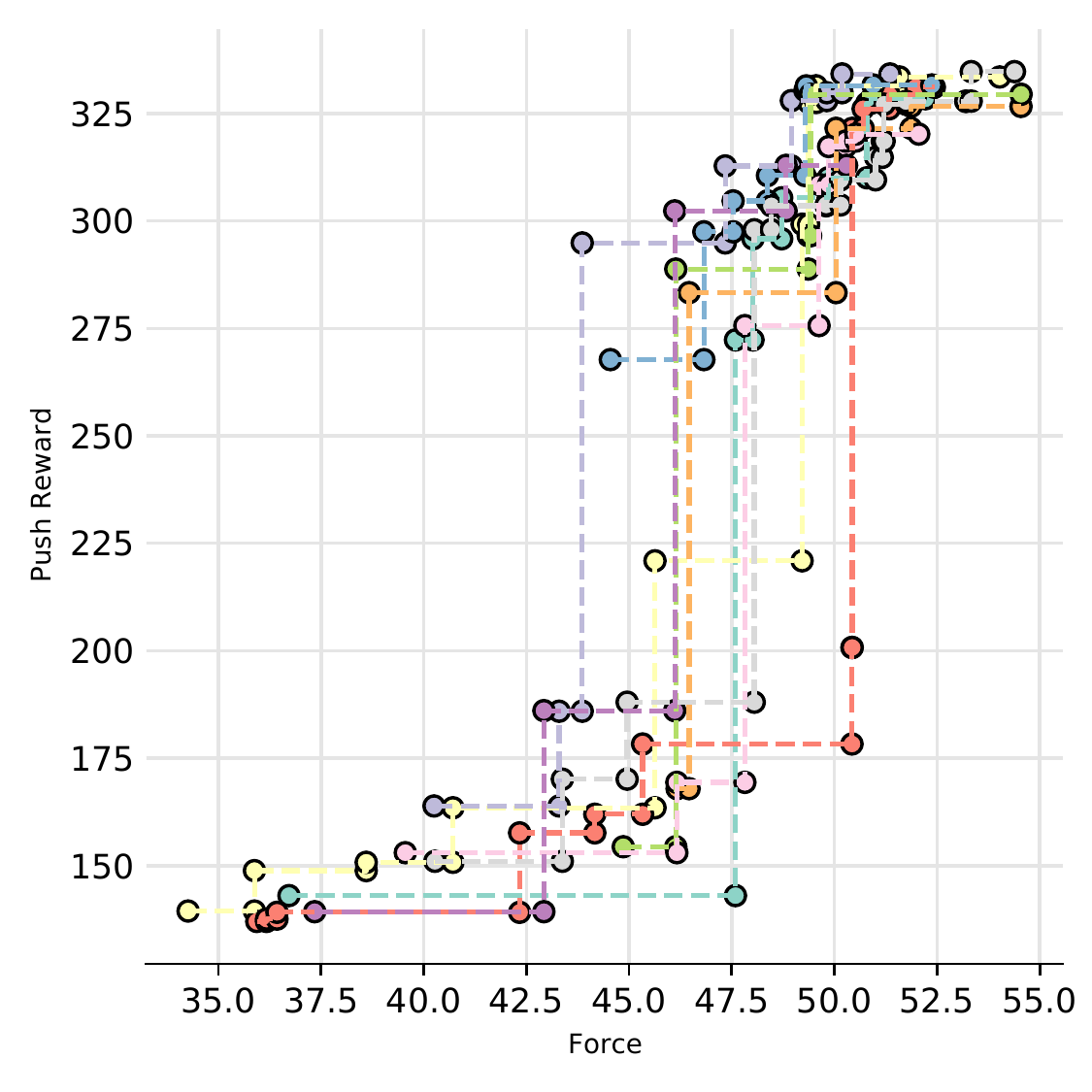}}}
	}
	\caption{Pareto front of the push task. Each experiment has a different color and each point represents a Pareto-optimal solution. It shows that higher rewards for pushing require higher interaction forces with the environment.}
	\label{fig:push-task-pareto}
	\vspace{-1.4em}
\end{figure}
We learn for 400 iterations and repeat the experiment 10 times. In order to obtain solutions that are robust enough to translate to the real system, we apply domain randomization. Each parameter set is evaluated in 7 worlds. Each execution uniformly samples one out of the four start positions for the robot arm. Furthermore, we vary the location of the object and the goal in the horizontal $(x)$ and $(y)$ directions by sampling from a Gaussian distribution with a standard deviation of \SI{7}{\milli\meter}.

We compare the learned solutions with (a) the outcome of a direct planner solution without any offset on the start and goal pose while pushing, (b) ten sets of random parameters from the search space and (c) the policies that are parameterized by the robot operators. We evaluated on the four start configurations used for learning as well as on two additional unknown ones. The results are shown in~Fig.~\ref{fig:success-results}a.

The results of a multi-objective optimization are parameters found along a Pareto front (Sec.~\ref{sec:multi-objective}, see Fig.~\ref{fig:push-task-pareto}). It contained \num{8.3} points on average, of which some minimize the impact on the environment to an extent that the push is not successful. An operator can choose a solution that is a good compromise between the success of the task on the real system and the force applied on the environment. The performance of one of the solutions that existed on the Pareto front is shown in Fig.~\ref{fig:success-results}.

Furthermore, we asked six robot operators to find values for the learnable parameters of the skill sequences. They were given the same start positions used for learning and were given a script to reset the arm to a start position of their choice. They could experiment with the system until they decided that their parameter set fulfills the criteria.
Their final parameter set that was also evaluated on the known and unknown start configurations.
On average the operators spent \SI{16.3(64)}{\minute} and executed \num{11.1(30)} trials on the system to configure this task.
Four out of the six operators found solutions that achieved the task from every start state. However, two of the operators' final parameters only achieved success rates of \SI{50}{\percent} and \SI{16.66}{\percent}.

\subsection{Peg-in-Hole Task}
\label{sec:peg-in-hole-task}
The PDDL goal of the peg insertion task is \texttt{(skiros:at skiros:Peg-1 skiros:BoxWithHole-1)}. The BT that is generated by the planner is shown in Fig.~\ref{fig:behavior_tree} and uses two skills: 1) $GoToLinear$ skill and 2) $PegInsertion$ skill. The first skill moves the end effector from its current location to the \emph{approach pose} of the hole. Once it is reached, the peg insertion procedure starts.

The $PegInsertion$ skill starts when the end effector reaches the approach pose of the box. It uses four separate \textit{SkiROS} primitive skills to 1) set the stiffness of the end effector to zero in (\(z\)) direction, 2) apply a downward force in (\(z\)) direction, 3) configure the center of the box as a goal and 4) additionally apply an overlaying circular search motion on top of the reference pose of the end effector as described in~\cite{mayr21iros}. The BT returns success only if the peg is inserted into the box hole by more than \SI{0.01}{\meter}.

We model the peg insertion as a multi-objective and multi-reward task. There are two objectives of the task, 1) successful insertion and 2) applied force. To assess the efficacy of the first objective, we use three rewards, 1) success of the BT, 2) peg distance to the hole, and 3) peg distance to the box. For the second objective, we use a single reward that measures the total force applied by the peg. There are three learnable parameters in this task, 1) downward force applied by the robot arm, 2) radius of the overlay search motion and 3) path velocity of the overlay search motion.  

\begin{figure}[tpb]
	{
		\setlength{\fboxrule}{0pt}
		\framebox{\parbox{3in}{
			\includegraphics[width=0.95\columnwidth]{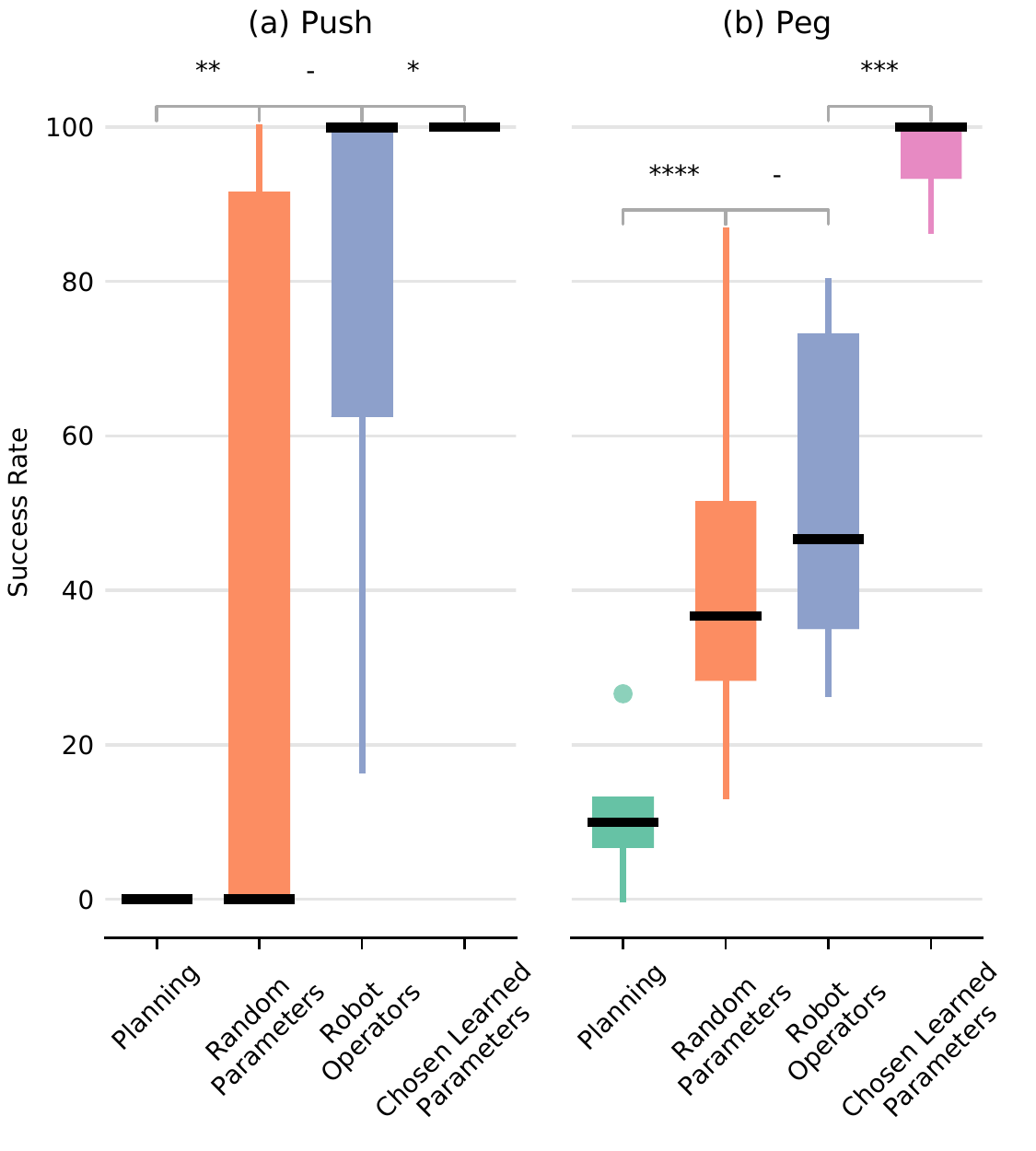}}}
	}
	\caption{The success rates of both experiments. The box plots show the median (black line) and interquartile range (25\(^{th}\) and 75\(^{th}\) percentile); the lines extend to the most extreme data points not considered outliers, and outliers are plotted individually. The number of stars indicates that the p-value of the Mann-Whitney U test is less than 0.1, 0.05, 0.01 and 0.001 respectively.}
	\label{fig:success-results}
	\vspace{-1.5em}
\end{figure}

We learn for 400 iterations in the simulation and repeat this experiment 10 times. To increase the robustness of the solutions we use domain randomization and evaluate each parameter configuration in 7 worlds. We vary the location of the box by sampling from a Gaussian distribution with a standard deviation of \SI{7}{\milli\meter} and uniformly sample one out of 5 start configurations of the robot arm. 
We compare the performance of the learned policies with (1) the outcome of the planner without a parameterized search motion, (2) randomly chosen parameter configurations from the parameter search space used for learning and (3) policies that are parameterized by human operators (see Fig.~\ref{fig:success-results}b).

The learned Pareto-optimal configurations consist of \num{6.1} points on average. We evaluated the insertion success using the 5 known and additional 10 unknown start configurations of the robot (Fig.~\ref{fig:success-results}b). 

To find policies for this task, the human operators took \SI{31.8(109)}{\minute} and executed \num{39(14)} trials on the system. However, compared to the randomly sampled policies the average insertion rate only increased from \SI{41}{\percent} to \SI{52.2}{\percent}. This is much lower than the average insertion rate of \SI{96}{\percent} of the best learned policies as shown in box four, Fig.~\ref{fig:success-results}b.
Furthermore, the average force that was chosen by the operators compared to the learned policies was \SI{16.6}{\percent} higher. Finally, the successful insertions by the learned policies were also \SI{18.1}{\percent} faster. Therefore, the learned policies outperformed the human operators in both objectives while also producing more reliable results.
%
%
\begin{figure}[tpb]
	{
		\setlength{\fboxrule}{0pt}
		\framebox{\parbox{3in}{
		\includegraphics[width=0.9\columnwidth]{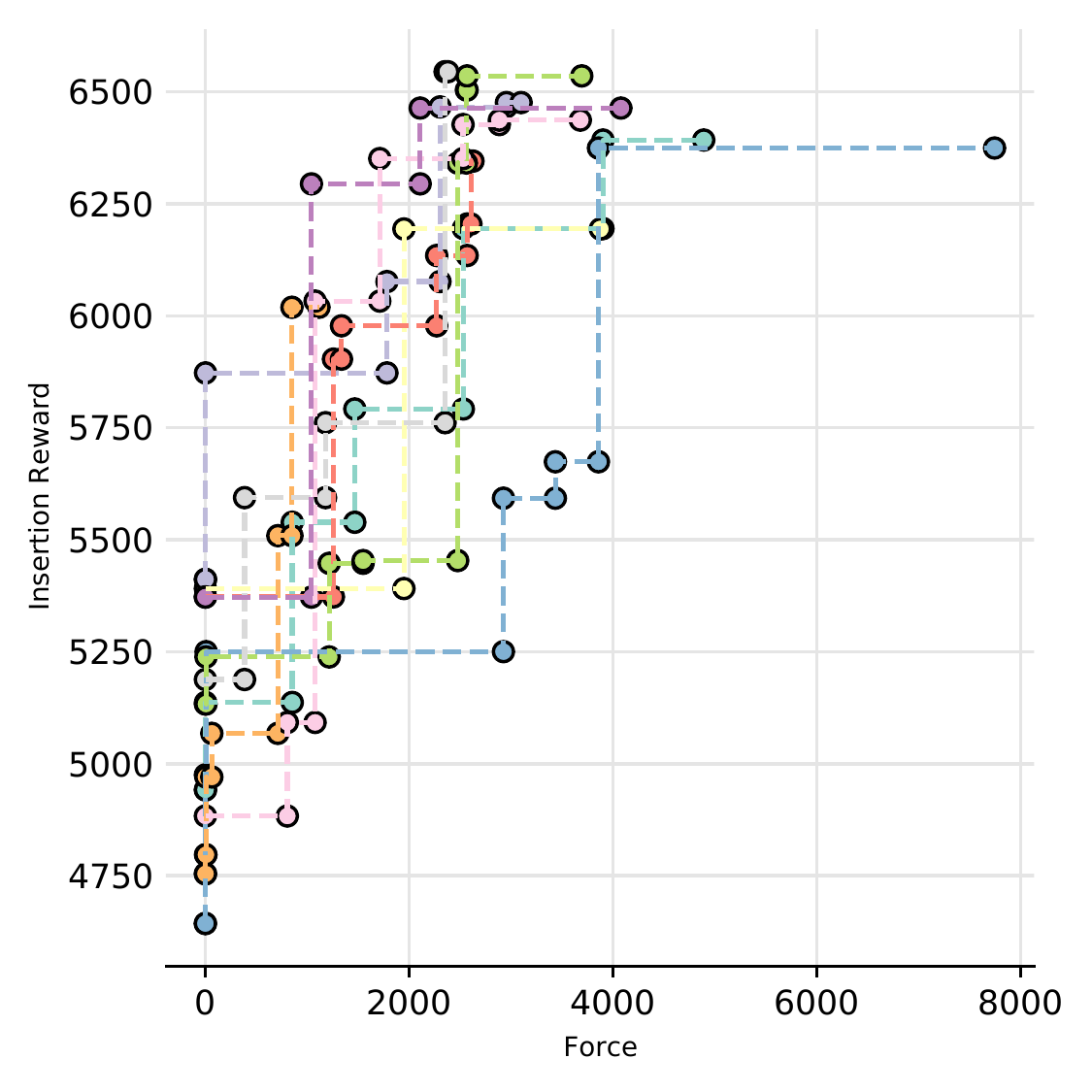}}}
	}
	\caption{Pareto front of the peg task. Each experiment has a different color. The goal is to maximize insertion reward while minimizing the interaction forces.}
	\label{fig:peg-in-hole-pareto}
	\vspace{-1.4em}
\end{figure}
\section{Conclusion}
In this paper we proposed a method for effectively combining task-level planning with learning to solve industrial contact-rich tasks. Our method leverages prior information and planning to acquire \emph{explicit} knowledge about the task, whereas it utilizes learning to capture the \emph{tacit} knowledge, i.e., the knowledge that is hard to formalize and which can only be captured through actual interaction.\todo{M: Add sth like "It therefore effectively combines deductive and inductive methods to solve robot tasks."}
We utilize behavior trees as an interpretable policy representation that is suitable for learning and leverage domain randomization for learning in simulation.
Finally, we formulate a multi-objective optimization scheme so that (1) we handle conflicting rewards adequately, and (2) an operator can choose a policy from the Pareto front and thus actively participate in the learning process.
%

We evaluated our method on two scenarios using a real \textit{KUKA} 7-DOF manipulator: (a) a pushing task, and (b) a peg insertion task. Both tasks are contact-rich and na\"{i}ve planning fails to solve them. The approach was able to outperform the baselines including the manual parameterization by robot operators.

For future work we are looking into multi-fidelity learning that can leverage a small amount of executions on the real system to complement the learning in simulation. Furthermore, the use of parameter priors for the optimum seems a promising direction to guide the policy search and make it more efficient.
%
%
\addtolength{\textheight}{-0.1cm}   
\section*{Appendix}
The implementation and the supplemental video are available at:\\\small{\url{https://sites.google.com/ulund.org/SkiREIL}}

\section*{Acknowledgement}
We thank Alexander Durr, Elin Anna Topp, Francesco Rovida and Jacek Malec for the interesting discussions and the constructive feedback.

This work was partially supported by the Wallenberg AI, Autonomous Systems and Software Program (WASP) funded by Knut and Alice Wallenberg Foundation.
This research was also supported in part by affiliate members and other supporters of the Stanford DAWN project—Ant Financial, Facebook, Google, InfoSys, Teradata, NEC, and VMware.


\bibliography{root}
\bibliographystyle{bib/IEEEtran}

\end{document}